\documentclass[10pt,twocolumn,letterpaper]{article}

\usepackage[pagenumbers]{cvpr}
\usepackage{times}
\usepackage{epsfig}
\usepackage{graphicx}
\usepackage{amsmath,amssymb,mathtools}
\usepackage{booktabs}
\usepackage{multirow}
\usepackage{makecell}
\usepackage{tabularx}
\usepackage{microtype}
\usepackage{xcolor}
\usepackage{colortbl}
\usepackage{enumitem}
\usepackage{pifont}
\newcommand{\cmark}{\ding{51}}
\newcommand{\xmark}{\ding{55}}

\definecolor{tablegroup}{HTML}{EAF0F5}
\definecolor{tableours}{HTML}{F3F7FA}
\definecolor{tablehuman}{HTML}{E7F2EA}
\definecolor{tablebest}{HTML}{D7E9F5}
\definecolor{tablesecond}{HTML}{F8E5D2}
\definecolor{tablesplit}{HTML}{BCC8D1}
\newcolumntype{E}{>{\centering\arraybackslash}m{3.2em}}

\setlist[itemize]{leftmargin=*,nosep}
\renewcommand{\arraystretch}{1.08}

\definecolor{cvprblue}{rgb}{0.21,0.49,0.74}
\usepackage[pagebackref,breaklinks,colorlinks,allcolors=cvprblue]{hyperref}

\def\paperID{*****}
\def\confName{CVPR}
\def\confYear{2027}

\title{NavJev: Efficient Vision-Language Navigation via Action-Centric Visual\\ Compression and Discriminative Action-Semantic Memory}

\author{
Kai Sheng \quad Liuyi Wang$^{\dagger}$ \quad Jinlong Li \quad Haojie Dai\\
\quad Chengju Liu \quad Qijun Chen$^{\dagger}$ \\
College of Electronic and Information Engineering, Tongji University, Shanghai, China\\
{\tt\small \{2610859,wly,li\_jinlong,tju\_dhj,liuchengju,qjchen\}@tongji.edu.cn}
}

\begin{document}

\maketitle

\begingroup
\renewcommand{\thefootnote}{\fnsymbol{footnote}}
\footnotetext[2]{Corresponding authors.}
\endgroup

\begin{abstract}
Recent zero-shot Vision-and-Language Navigation (VLN) methods increasingly rely on multimodal large language models (MLLMs) to reason over visual observations, navigation instructions, and candidate actions. Although effective, repeatedly invoking autoregressive multimodal reasoning at every navigation step introduces substantial inference latency, limiting the responsiveness of embodied agents. We propose NavJev, an efficient VLN framework that reformulates online navigation from repeated multimodal generation into compact visual compression followed by lightweight typed action selection. Specifically, Action-Centric Visual Compression (ACVC) integrates waypoint geometry, BLIP captions, and RAM semantic tags into compact representations of candidate actions, while Discriminative Action-Semantic Memory (DASM) filters shared semantics and maintains discriminative action-specific evidence across navigation steps. Based on these representations, Jev directly performs structured probabilistic decisions over the available action set. Experiments on R2R-CE show that NavJev achieves 27.0\% SR and 22.4\% SPL with only 0.65\,s per navigation step, while substantially reducing inference latency and cost compared with MLLM-based VLN methods. The project page is available at \url{https://kai-sheng-caesar.github.io/NavJev/}.
\end{abstract}

\section{Introduction}
\label{sec:intro}

Vision-and-Language Navigation (VLN) requires an embodied agent to understand natural-language instructions and interact with visual environments to reach a target location~\cite{anderson2018r2r,wang2024goat,krantz2020vlnce,wang2025rethinking}. Recent Multimodal Large Language Models (MLLMs) have enabled a new class of zero-shot VLN methods that perform navigation without task-specific training~\cite{wang2026comprehensive}. By reasoning over panoramic observations, navigation instructions, history, and candidate waypoints, these methods show promising generalization to unseen environments~\cite{qiao2025opennav,shi2025smartway}. However, this capability comes with substantial computational overhead. Large multimodal models are repeatedly invoked at every navigation step, increasing online inference cost and limiting the real-time responsiveness required by embodied agents.

\begin{figure}[t]
\centering
\includegraphics[width=\columnwidth]{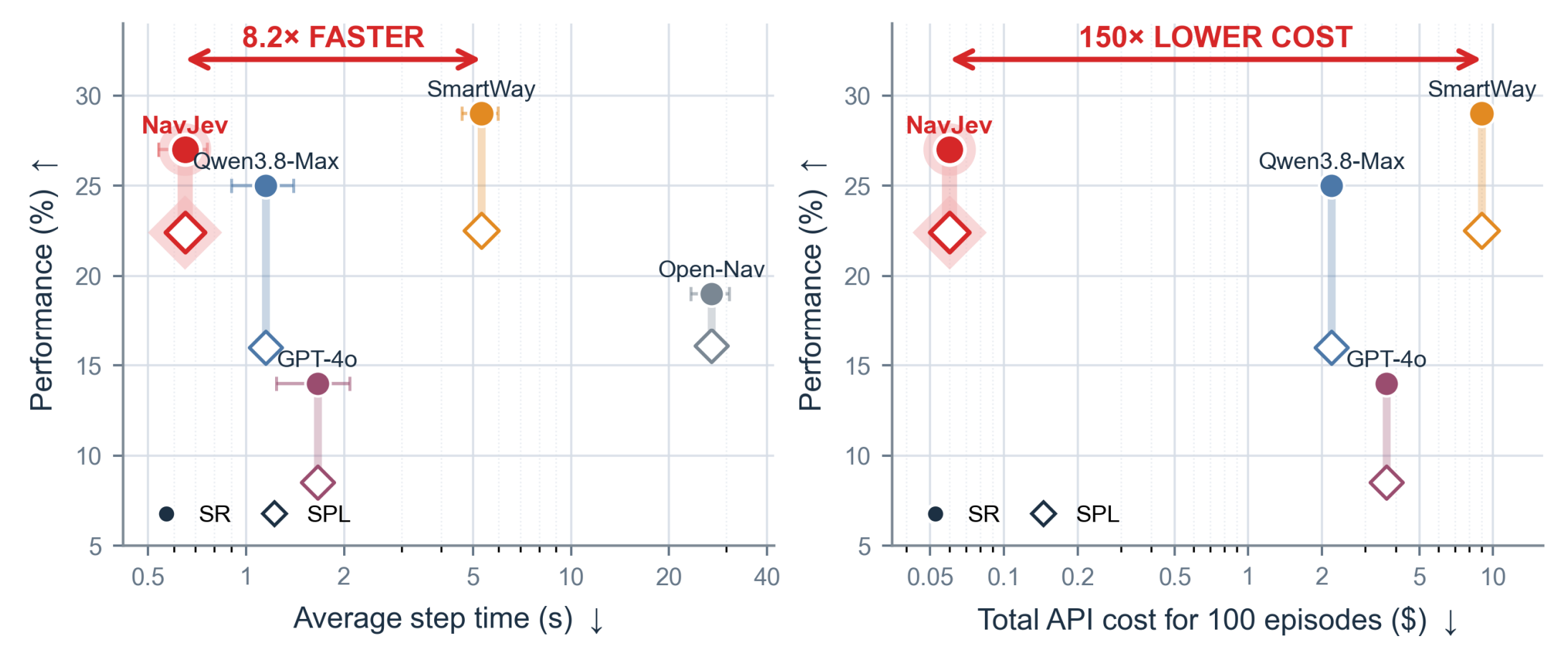}
\caption{Performance and efficiency advantages of NavJev on R2R-CE. NavJev achieves 27.0\% SR and 22.4\% SPL with only 0.65\,s per navigation step and substantially lower inference cost than representative MLLM-based VLN methods.}
\label{fig:head}
\end{figure}

This efficiency issue is particularly evident because the final decision at each VLN step is inherently structured. Although an agent may observe a rich visual panorama, it ultimately only needs to select one waypoint from a small set of navigable candidates or decide to stop. Existing MLLM-based methods nevertheless repeatedly encode high-dimensional visual observations and autoregressively generate textual reasoning before obtaining the action decision, creating a mismatch between the complexity of multimodal inference and the structured nature of navigation decisions.

Recently, Jev introduced a different decision paradigm~\cite{typesafe2026jev}. Rather than generating open-ended text token by token, Jev directly maps a given state to typed probabilistic decisions. Specifically, its Choice type selects among predefined alternatives and outputs a probability distribution without autoregressive text generation. This property is naturally aligned with VLN: candidate waypoints already define a finite action set, and navigation requires repeatedly selecting the action that best matches the current instruction and state. Jev currently operates on textual or structured states, which provide a substantially more compact interface for decision making than raw visual observations. This raises a key question for VLN: how can rich visual observations be compressed into compact textual representations while preserving the discriminative information required for navigation decisions?

To address this question, we propose NavJev, an efficient vision-language navigation framework that bridges visual observations and typed decision making through compact action-centric representations. Instead of repeatedly sending images to an MLLM, NavJev converts the local visual environment into textual representations centered on executable navigation actions. Specifically, Action-Centric Visual Compression (ACVC) associates each candidate waypoint with its relative direction and distance, a BLIP caption describing the corresponding visual observation, and RAM semantic tags identifying salient objects and scene concepts. Together, these cues compress raw visual observations into action descriptions of where the agent can go and what lies along each direction, providing a compact interface for subsequent typed action selection.

However, visual compression alone is insufficient because neighboring candidate directions often share substantial scene semantics. Objects or scene attributes shared across candidate actions increase representation redundancy while providing little information for distinguishing among them. We therefore introduce Discriminative Action-Semantic Memory (DASM), which filters shared semantics and maintains action-specific semantic evidence across navigation steps. By organizing the discriminative semantics of current candidate actions together with historical action-semantic information, DASM provides Jev with a compact structured state for sequential navigation decision making.

Based on these representations, NavJev formulates each navigation step as a typed selection problem over candidate actions. Given the navigation instruction, history, and discriminative action-semantic state, Jev directly evaluates the available actions and selects the next waypoint through typed probabilistic decisions. In this way, NavJev decouples visual understanding from online action selection: ACVC compresses local observations into action-centric representations, DASM maintains discriminative action-semantic information across navigation steps, and Jev efficiently performs continuous closed-loop decisions over the resulting structured state.

Experiments on R2R-CE show that this paradigm substantially improves navigation efficiency while maintaining competitive zero-shot performance. As shown in Fig.~\ref{fig:head}, NavJev achieves 27.0\% SR and 22.4\% SPL with an average inference time of only 0.65\,s per navigation step, while substantially reducing inference latency and cost compared with representative MLLM-based VLN methods. Real-world experiments in office and caf\'e environments further show that this efficiency advantage transfers to physical navigation. These results indicate that effective zero-shot VLN does not necessarily require expensive multimodal generation at every step, and that action-centric representations, discriminative action-semantic memory, and typed decision making provide an efficient alternative for embodied navigation.

Our contributions are threefold:
\begin{itemize}
\item We propose NavJev, an efficient zero-shot VLN framework that replaces repeated online multimodal generation with compact action-centric representations and lightweight typed action selection.

\item We introduce Action-Centric Visual Compression (ACVC) and Discriminative Action-Semantic Memory (DASM), which construct compact action representations and preserve action-specific semantics across navigation steps.

\item Experiments on R2R-CE show that NavJev achieves 27.0\% SR and 22.4\% SPL with only 0.65\,s per navigation step, substantially reducing inference latency and cost compared with representative MLLM-based VLN methods.
\end{itemize}
\section{Related Work}
\label{sec:related}

\paragraph{Vision-Language Navigation.}
Vision-Language Navigation (VLN) requires an embodied agent to follow natural-language instructions and navigate through indoor environments~\cite{anderson2018r2r,wang2026magic}, while VLN-CE further extended the task to continuous environments without predefined navigation graphs~\cite{krantz2020vlnce,an2024etpnav}. Early methods mainly improved navigation through vision-language pretraining and history-aware reasoning, such as PREVALENT~\cite{hao2020prevalent}, VLN$\circlearrowright$BERT~\cite{hong2021vlnbert}, HAMT~\cite{chen2021hamt}, and DUET~\cite{chen2022duet}. Subsequent work further explored scalable data generation, spatial representations, and continuous waypoint planning, including ScaleVLN~\cite{wang2023scalevln}, BEVBert~\cite{an2023bevbert}, and ETPNav~\cite{an2024etpnav}. Recent advances in multimodal large language models have reshaped the navigation paradigm~\cite{wang2026comprehensive}. Uni-NaViD unifies multiple navigation tasks through a shared video--action interface~\cite{zhang2024uninavid}. StreamVLN maintains bounded long-horizon context with slow--fast memory~\cite{wei2025streamvln}. JanusVLN decouples semantic and spatial memory~\cite{zeng2025janusvln}, while CLASH coordinates large and small models through a hierarchical navigation framework~\cite{wang2025clash}. These methods mainly focus on improving navigation accuracy and generalization, whereas NavJev places greater emphasis on reducing the online inference overhead of repeated closed-loop decision making.

\begin{figure*}[t]
    \centering
    \includegraphics[width=\textwidth]{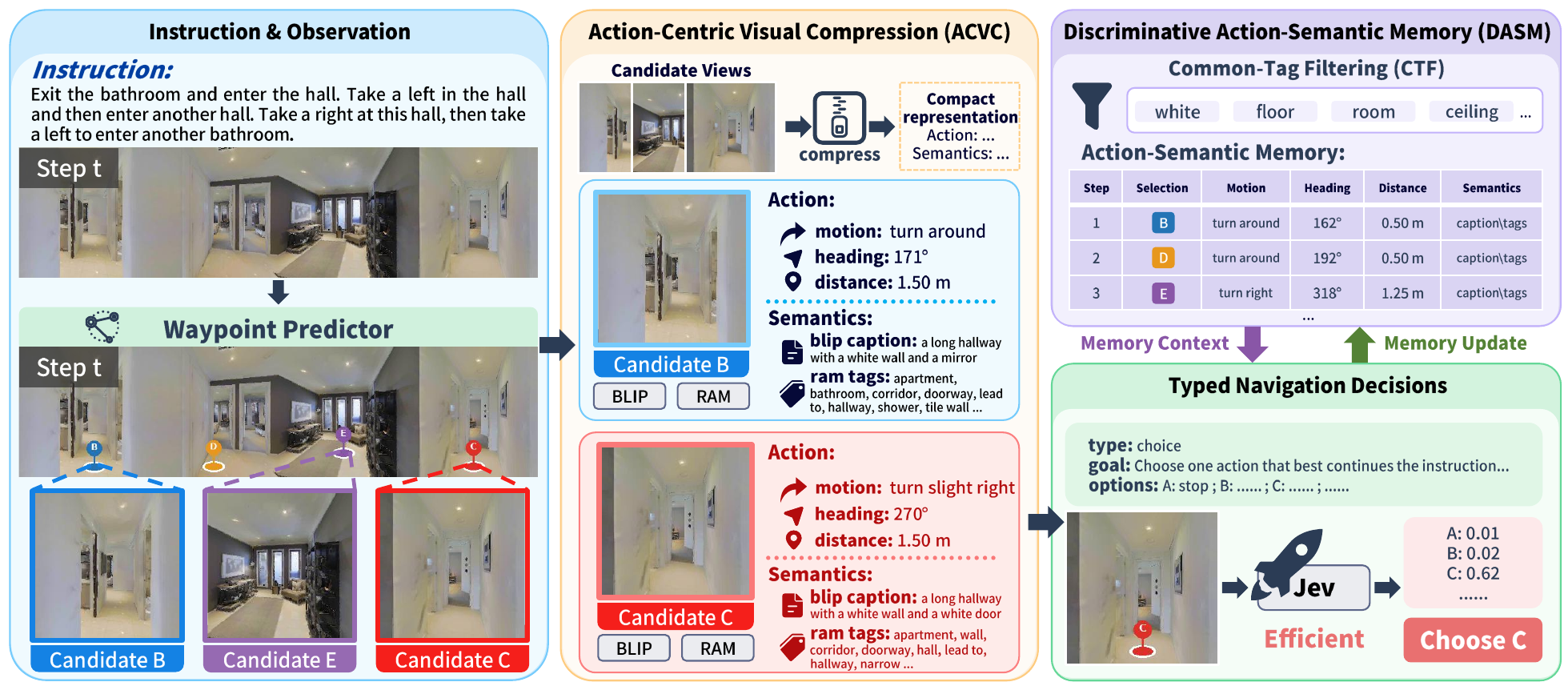}
    \caption{Overview of NavJev. ACVC compresses candidate observations into compact action-centric representations, DASM filters shared semantics and maintains discriminative action-specific evidence across navigation steps, and Jev performs efficient typed action selection.}
    \label{fig:navjev}
\end{figure*}

\paragraph{Large Models for Navigation.}
Recent advances in multimodal large language models (MLLMs) have promoted zero-shot navigation through online reasoning and semantic understanding~\cite{bai2025qwen3,achiam2023gpt}. DiscussNav performs action selection through multi-expert discussion~\cite{long2023discuss}, NavGPT uses LLM-based reasoning for navigation~\cite{zhou2024navgpt}, and MapGPT combines map-guided prompting with adaptive path planning~\cite{chen2024mapgpt}. Open-Nav performs spatial-temporal reasoning with open-source large models~\cite{qiao2025opennav}, while SmartWay improves waypoint prediction and backtracking~\cite{shi2025smartway}. Recent methods further structure the large-model decision process. AO-Planner introduces affordance-oriented planning~\cite{chen2025aoplanner}, and P2DNav decomposes panoramic direction reasoning from local down-view grounding~\cite{sheng2026p2dnav}. Despite their strong zero-shot performance, these methods still rely on repeated autoregressive large-model inference during navigation. In contrast, NavJev reformulates navigation as typed selection over structured action states for efficient decision making.

\section{Preliminaries}
\label{sec:preliminaries}

We consider Vision-and-Language Navigation in Continuous Environments (VLN-CE)~\cite{krantz2020vlnce}, where an embodied agent follows a natural-language instruction $\mathcal{L}$ to navigate to a target location in a continuous 3D environment. At navigation step $t$, the agent receives a panoramic RGB-D observation
\begin{equation}
\mathcal{I}_t=
\left\{
(I^{\mathrm{rgb}}_{t,i},
I^{\mathrm{depth}}_{t,i})
\right\}_{i=1}^{K},
\end{equation}
where $K=12$ views uniformly cover $360^\circ$. Based on the current observation, the agent constructs a set of navigable waypoint candidates
$\mathcal{W}_t=\{w_t^1,\ldots,w_t^{M_t}\}$, where each waypoint specifies a local navigation direction and distance. Together with the termination action, the action space is defined as
\begin{equation}
\mathcal{A}_t=
\{w_t^1,\ldots,w_t^{M_t},\textsc{Stop}\}.
\end{equation}

Given the instruction $\mathcal{L}$, navigation history $\mathcal{H}_t$, and current observation $\mathcal{I}_t$, the agent selects an action $a_t\in\mathcal{A}_t$ until issuing \textsc{Stop} or reaching the maximum navigation horizon. The objective is to reach and stop near the target while maintaining an efficient navigation trajectory.
\section{Method}
\label{sec:method}

\subsection{Overview}
\label{sec:method_overview}

We introduce NavJev, an efficient zero-shot vision-and-language navigation (VLN) framework that replaces repeated multimodal generation with compact action-centric representations, discriminative action-semantic memory, and typed decisions. The overall architecture is illustrated in Fig.~\ref{fig:navjev}. Action-Centric Visual Compression (ACVC) converts candidate waypoint observations into compact textual action representations by integrating navigation geometry, BLIP captions, and RAM semantic tags (Section~\ref{sec:acvc}). Discriminative Action-Semantic Memory (DASM) filters shared semantics and maintains action-specific semantic evidence across navigation steps (Section~\ref{sec:dasm}). Finally, Jev performs typed selection over the available navigation actions through a constrained option-label interface (Section~\ref{sec:typed_decision}).

\subsection{Action-Centric Visual Compression}
\label{sec:acvc}

NavJev first constructs a set of executable waypoint candidates from the panoramic RGB-D observation. We retain the pretrained waypoint predictor of SmartWay~\cite{shi2025smartway}. Given 12 RGB-D views sampled at $30^{\circ}$ intervals, DINOv2~\cite{oquab2024dinov2} extracts RGB features, while a depth encoder captures local geometric information. The two modalities are then fused through cross-attention and processed by a Transformer waypoint predictor to estimate a navigability heatmap:
\begin{equation}
\mathbf{P}_t =
f_{\mathrm{wp}}\!\left(
f_{\mathrm{rgb}}(\mathcal{I}_t^{\mathrm{rgb}}),
f_{\mathrm{depth}}(\mathcal{I}_t^{\mathrm{depth}})
\right).
\end{equation}

The resulting heatmap $\mathbf{P}_t$ contains 120 heading bins with a $3^{\circ}$ resolution and 12 distance bins separated by $0.25\,\mathrm{m}$. Non-maximum suppression is then applied to extract up to five navigable waypoint candidates. Each resulting waypoint is represented as
\begin{equation}
w_t^i =
\left(
\theta_t^i,
d_t^i,
I_t^i
\right),
\end{equation}
where $\theta_t^i$ and $d_t^i$ denote the relative heading and distance of the candidate, respectively, and $I_t^i$ denotes the RGB sub-view closest to the predicted heading. Subsequent visual processing is therefore restricted to action-relevant views.

Directly sending these candidate images to a multimodal large model incurs substantial visual encoding and generation overhead. We therefore introduce Action-Centric Visual Compression (ACVC), which converts each candidate observation into a compact textual representation centered on an executable navigation action. For waypoint $w_t^i$, BLIP~\cite{li2022blip} generates a short visual caption $C_t^i$, while RAM~\cite{zhang2024ram} extracts a set of semantic tags$R_t^i$:
\begin{equation}
C_t^i = f_{\mathrm{BLIP}}(I_t^i),
\qquad
R_t^i = f_{\mathrm{RAM}}(I_t^i).
\end{equation}
These two forms of visual information are complementary. The BLIP caption provides a concise description of the local scene and spatial context, while RAM explicitly identifies salient objects and scene concepts.

We further associate these visual cues with the corresponding waypoint geometry and motion description $m_t^i$, such as \emph{go forward}, \emph{turn slight left}, or \emph{turn sharp right}. The resulting action-centric representation is
\begin{equation}
Z_t^i =
\left(
m_t^i,
\theta_t^i,
d_t^i,
C_t^i,
R_t^i
\right).
\end{equation}

In this way, ACVC directly binds geometric and semantic information to the corresponding executable action. The representation describes both where the agent can move and what can be observed along that direction. The complete compressed candidate set is defined as
\begin{equation}
\mathcal{Z}_t =
\left\{
Z_t^i
\right\}_{i=1}^{M_t}.
\end{equation}
This allows subsequent navigation decisions to operate directly on compact textual action representations rather than raw candidate images.

ACVC is also optimized for low-latency perception. Candidate images are processed in batches. Since neighboring waypoint candidates may map to the same panoramic view, duplicate images are processed only once and the resulting outputs are reused across the corresponding actions. In addition, BLIP caption and RAM tagging generation operate on the same image batch and can be executed in parallel. The effective ACVC latency can therefore be approximated as
\begin{equation}
T_{\mathrm{ACVC}}
\approx
\max\!\left(
T_{\mathrm{BLIP}},
T_{\mathrm{RAM}}
\right)
+
\epsilon,
\end{equation}
where $\epsilon$ denotes the additional scheduling and synchronization overhead.

\subsection{Discriminative Action-Semantic Memory}
\label{sec:dasm}

Although ACVC produces compact action representations, neighboring waypoint candidates often contain substantial semantic overlap because they correspond to nearby regions of the same scene. Repeated objects and scene attributes increase textual redundancy while providing limited information for distinguishing which action better matches the navigation instruction. We therefore introduce Discriminative Action-Semantic Memory (DASM), which combines intra-step semantic filtering with inter-step retention of action-specific semantic evidence.

Given the RAM tag sets $\{R_t^i\}_{i=1}^{M_t}$ associated with the current waypoint candidates, DASM first performs Common-Tag Filtering (CTF) to identify and remove semantics shared across all candidate directions:
\begin{equation}
G_t =
\bigcap_{i=1}^{M_t} R_t^i.
\end{equation}
The shared set $G_t$ mainly contains scene information visible across all candidate directions. Rather than repeatedly attaching these tags to each action, DASM retains them once as shared scene context and removes them from the action-specific tag sets:
\begin{equation}
\widetilde{R}_t^i =
R_t^i \setminus G_t,
\qquad
i=1,\ldots,M_t.
\end{equation}

The discriminative representation of each candidate action is then defined as
\begin{equation}
D_t^i =
\left(
m_t^i,
\theta_t^i,
d_t^i,
C_t^i,
\widetilde{R}_t^i
\right),
\qquad
i=1,\ldots,M_t.
\end{equation}
Compared with the original ACVC representation $Z_t^i$, $D_t^i$ emphasizes semantic differences among candidate actions while avoiding repeated scene-level information.

DASM further incorporates the semantic evidence associated with each executed action into the navigation history $\mathcal{H}_t$. After action $a_t$ is executed, the history is updated as
\begin{equation}
\mathcal{H}_{t+1}
=
\mathcal{H}_t
\oplus
\phi(a_t),
\end{equation}
where $\oplus$ denotes ordered append and $\phi(a_t)$ stores the executed action together with its motion description, heading, distance, BLIP caption, and filtered RAM tags. In this way, the navigation history records not only the agent's executed movements, but also the discriminative visual evidence associated with previous decisions.

The structured action-semantic state provided to the decision module is defined as
\begin{equation}
\mathcal{S}_t =
\left(
\mathcal{H}_t,
G_t,
\left\{
D_t^i
\right\}_{i=1}^{M_t}
\right).
\end{equation}
DASM therefore combines the historical semantics of executed actions with the discriminative evidence of current waypoint candidates, providing Jev with a compact temporal state for subsequent typed action selection.

\subsection{Typed Navigation Decision}
\label{sec:typed_decision}

Given the navigation instruction $\mathcal{L}$ and the structured action-semantic state $\mathcal{S}_t$, NavJev formulates each navigation step as a typed selection problem. Each available action in the current action space $\mathcal{A}_t$ is assigned a corresponding option label, and Jev directly selects an action:
\begin{equation}
a_t =
f_{\mathrm{Jev}}\!\left(
\mathcal{L},
\mathcal{S}_t,
\mathcal{A}_t
\right),
\qquad
a_t \in \mathcal{A}_t.
\end{equation}

The selected label is then mapped to the corresponding executable action. If the selected action corresponds to waypoint $w_t^j$, the continuous navigation command is
\begin{equation}
u_t =
\left(
\theta_t^j,
d_t^j
\right),
\qquad
a_t = w_t^j,
\end{equation}
which is passed to the underlying continuous controller. Selecting \textsc{Stop} terminates the episode.

After each nonterminal action execution, the executed action and its associated semantic evidence are incorporated into $\mathcal{H}_t$, the local observation is updated, and the ACVC--DASM--Jev pipeline is repeated. Through this closed-loop process, NavJev replaces repeated multimodal generation with compact action-centric representations, discriminative action-semantic memory, and typed action selection.

\begin{figure*}[t]
    \centering
    \includegraphics[width=0.85\textwidth]{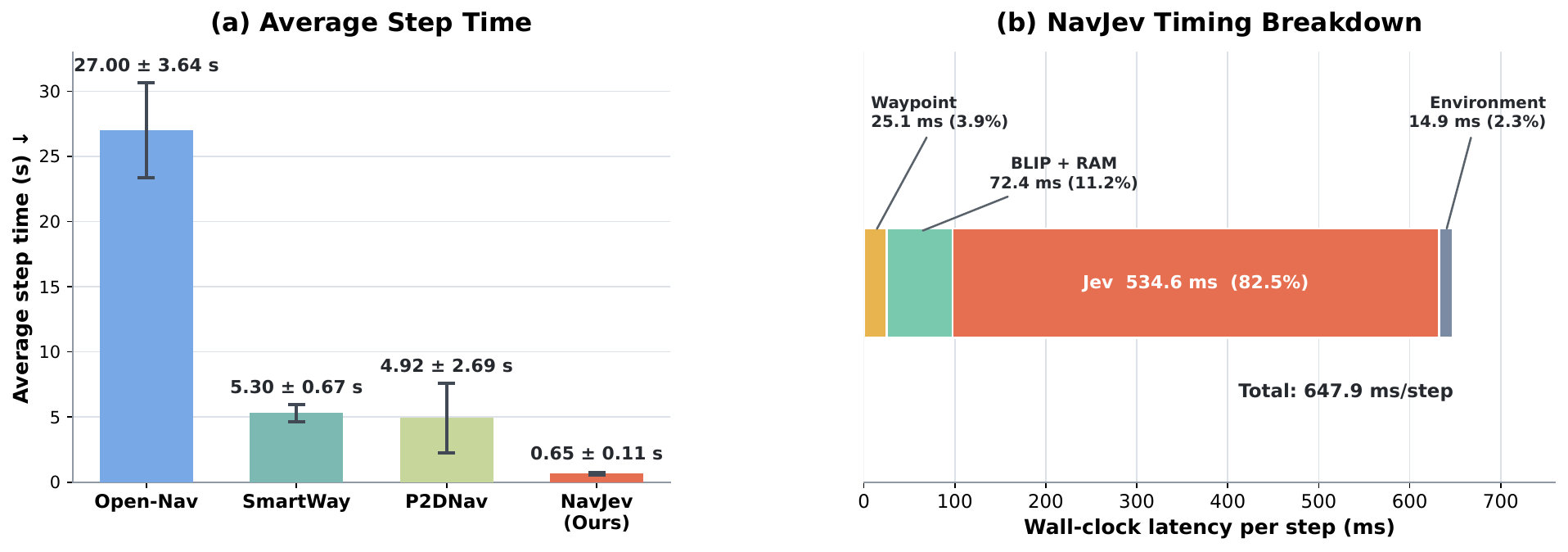}
    \caption{Inference efficiency analysis of NavJev. (a) Average step time compared with representative zero-shot VLN methods. NavJev requires only 0.65\,s per step, achieving approximately 41.5$\times$, 8.2$\times$, and 7.6$\times$ speedups over Open-Nav, SmartWay, and P2DNav, respectively. (b) Wall-clock latency breakdown of NavJev. Jev API calls account for most of the inference time, while waypoint prediction, ACVC perception, and other modules introduce only limited additional overhead.}
    \label{fig:timing_comparison}
\end{figure*}

\section{Experiments}
\label{sec:experiments}

\subsection{Experimental Setup}

\paragraph{Dataset and Evaluation Setting.}
Following recent zero-shot VLN methods~\cite{qiao2025opennav,shi2025smartway,sheng2026p2dnav}, we evaluate NavJev on 100 episodes from the R2R-CE val-unseen split under the Open-Nav evaluation framework~\cite{qiao2025opennav}. This setting enables direct comparison with MLLM-based navigation approaches under the same zero-shot protocol. Representative supervised methods evaluated on the full val-unseen split are also included as reference.

\begin{table}[t]
\centering
\caption{Comparison with representative supervised and zero-shot methods on the R2R-CE val-unseen split.}
\label{tab:r2r_ce}

\scriptsize
\setlength{\tabcolsep}{3.0pt}
\renewcommand{\arraystretch}{1.05}

\resizebox{1\columnwidth}{!}{%
\begin{tabular}{llcccc}
\toprule
\textbf{Method} &
\textbf{Decision Model} &
\textbf{NE}$\downarrow$ &
\textbf{OSR}$\uparrow$ &
\textbf{SR}$\uparrow$ &
\textbf{SPL}$\uparrow$ \\
\midrule

\multicolumn{6}{l}{\textit{Supervised Learning}} \\
\midrule
Seq2Seq~\cite{krantz2020vlnce}
& -- & 7.77 & 37.0 & 25.0 & 22.0 \\

MEE~\cite{mee}
& -- & 6.82 & 44.6 & 35.9 & 32.3 \\

NaVid~\cite{zhang2024navid}
& Vicuna-7B
& 5.47 & 49.1 & 37.4 & 35.9 \\

MLANet~\cite{he2025multilevel}
& -- & 6.30 & 42.0 & 38.0 & 35.0 \\


Uni-NaVid~\cite{zhang2024uninavid}
& Vicuna-7B
& 5.58 & 53.3 & 47.0 & 42.7 \\

NaVILA~\cite{cheng2025navila}
& Llama-3-8B
& 5.22 & 62.5 & 54.0 & 49.0 \\

StreamVLN~\cite{wei2025streamvln}
& Qwen2-7B
& 4.98 & 64.2 & 56.9 & 51.9 \\

ETPNav~\cite{an2024etpnav}
& -- & 4.71 & 65.0 & 57.0 & 49.0 \\

BEVBert~\cite{an2023bevbert}
& -- & 4.57 & 67.0 & 59.0 & 50.0 \\

JanusVLN~\cite{zeng2025janusvln}
& Qwen2.5-VL-7B
& 4.78 & 65.2 & 60.5 & \textbf{56.8} \\

NavFoM~\cite{NavFoM}
& Qwen2-7B
& 4.61 & 72.1 & 61.7 & 55.3 \\

CLASH~\cite{wang2025clash}
& Qwen2.5-VL-72B
& \textbf{4.06} & \textbf{73.0} & \textbf{65.0} & 55.0 \\


\midrule
\multicolumn{6}{l}{\textit{Zero-shot}} \\
\midrule
Random
& --
& 8.63 & 12.0 & 2.0 & 1.5 \\

MapGPT-CE~\cite{chen2024mapgpt}
& GPT-4o
& 8.16 & 21.0 & 7.0 & 5.0 \\

DiscussNav-CE~\cite{long2023discuss}
& GPT-4
& 7.77 & 15.0 & 11.0 & 10.5 \\

Open-Nav~\cite{qiao2025opennav}
& Llama3.1-70B
& 7.25 & 23.0 & 16.0 & 12.9 \\

Open-Nav~\cite{qiao2025opennav}
& GPT-4o
& 6.70 & 23.0 & 19.0 & 16.1 \\

CA-Nav~\cite{chen2025constraint}
& GPT-4
& 7.58 & 48.0 & 25.3 & 10.8 \\

SmartWay~\cite{shi2025smartway}
& GPT-4o
& 7.01 & 51.0 & 29.0 & 22.5 \\

P2DNav~\cite{sheng2026p2dnav}
& Qwen3-VL-32B
& \textbf{6.61} & \textbf{65.0} & \textbf{50.0} & \textbf{30.6} \\

\midrule
\textbf{NavJev (Ours)}
& \textbf{Jev}
& \textbf{7.48} & \textbf{35.0} & \textbf{27.0} & \textbf{22.4} \\

\bottomrule
\end{tabular}%
}
\end{table}

\paragraph{Metrics.}
We report standard VLN-CE metrics, including Navigation Error (NE), Oracle Success Rate (OSR), Success Rate (SR), and Success weighted by Path Length (SPL). NE measures the geodesic distance between the final agent position and the goal, while OSR measures whether the agent reaches the goal region at any point during navigation. SR evaluates successful termination within the goal region, and SPL further considers path efficiency. We additionally report normalized Dynamic Time Warping (nDTW) and Trajectory Length (TL) for trajectory-level analysis. For inference efficiency, we measure decision latency, average wall-clock time per navigation step, and API cost.

\paragraph{Implementation Details.}
All simulation experiments are conducted in Habitat on the R2R-CE val-unseen split. NavJev uses the pretrained waypoint predictor described in Sec.~\ref{sec:acvc}, and all local perception and decision-model inference is evaluated on the same NVIDIA RTX PRO 6000 GPU. The Jev decision model is accessed through its official API for typed action selection.
For real-world deployment, we use a wheeled robot equipped with an Insta360 X4 panoramic camera to acquire visual observations. Candidate waypoints are generated using the same clustering-based waypoint extraction strategy as CLASH~\cite{wang2025clash}, and the resulting candidates are processed by the same ACVC--DASM--Jev pipeline used in simulation. All server-side model inference is also performed on the NVIDIA RTX PRO 6000 GPU.

\subsection{Comparison with State-of-the-Art Methods}

Table~\ref{tab:r2r_ce} compares NavJev with representative supervised and zero-shot VLN methods. Zero-shot approaches, including DiscussNav-CE, MapGPT-CE, Open-Nav, SmartWay, and P2DNav, rely on MLLMs for navigation decisions. In contrast, NavJev replaces repeated multimodal generation with compact textual action representations and typed decisions.

NavJev achieves 27.0\% SR and 22.4\% SPL. Compared with Open-Nav using GPT-4o, NavJev improves SR from 19.0\% to 27.0\% and SPL from 16.1\% to 22.4\%, corresponding to gains of 8.0\% and 6.3\%, respectively. Compared with the Llama3.1-70B variant of Open-Nav, the gains increase to 11.0\% and 9.5\% in SR and SPL, respectively. NavJev achieves 22.4\% SPL, with performance comparable to SmartWay, while avoiding multimodal reasoning. These results show that compact action-centric representations and typed decisions can retain competitive zero-shot navigation performance with a substantially lighter online inference process.

\begin{table}[t]
\centering
\caption{Inference efficiency comparison with representative task-fine-tuned MLLM-based VLN methods. These methods typically predict low-level discrete actions at each step, whereas NavJev performs waypoint-level action selection.}
\label{tab:official_efficiency_compact}
\setlength{\tabcolsep}{4.0pt}
\resizebox{\columnwidth}{!}{%
\begin{tabular}{lcccc}
\toprule
\raisebox{0.8ex}{\textbf{Method}} &
\shortstack{\textbf{Average Step}\\\textbf{Time (s)}$\downarrow$} &
\shortstack{\textbf{Average Episode}\\\textbf{Time (s)}$\downarrow$} &
\shortstack{\textbf{Average Steps}\\\textbf{per Episode}} &
\shortstack{\textbf{Peak GPU}\\\textbf{Memory (GiB)}$\downarrow$} \\
\midrule
NaVid     & 0.28 & 24.60 & 89.2  & 17.95 \\
NaVILA    & 0.21 & 24.30 & 118.7 & 17.46 \\
StreamVLN & 0.23 & 16.69 & 73.6  & 23.63 \\
JanusVLN  & 0.64 & 44.22 & 68.9  & 35.80 \\
\midrule
NavJev    & 0.65 & \textbf{6.03} & \textbf{9.3} & \textbf{4.46} \\
\bottomrule
\end{tabular}%
}
\end{table}

\begin{table}[t]
\centering
\caption{Comparison of decision models under the same ACVC--DASM representation on the R2R-CE 100-episode split. Qwen3.8-Max and GPT-4o are evaluated with reasoning disabled.}
\label{tab:backend_comparison}

\setlength{\tabcolsep}{2.8pt}
\resizebox{\columnwidth}{!}{%
\begin{tabular}{lccccccc}
\toprule
\raisebox{0.8ex}{\textbf{Decision Model}} &
\raisebox{0.8ex}{\textbf{SR}$\uparrow$} &
\raisebox{0.8ex}{\textbf{SPL}$\uparrow$} &
\raisebox{0.8ex}{\textbf{OSR}$\uparrow$} &
\raisebox{0.8ex}{\textbf{nDTW}$\uparrow$} &
\shortstack{\textbf{Decision}\\\textbf{Latency (s)}$\downarrow$} &
\shortstack{\textbf{Average Step}\\\textbf{Time (s)}$\downarrow$} &
\shortstack{\textbf{Total}\\\textbf{Cost}$\downarrow$} \\
\midrule

GPT-4o
& 14.0
& 8.5
& 42.0
& 23.8
& 1.54 $\pm$ 0.42
& 1.66 $\pm$ 0.42
& \$3.67 \\

Qwen3.8-Max
& 25.0
& 16.0
& \textbf{46.0}
& 34.2
& 1.03 $\pm$ 0.25
& 1.15 $\pm$ 0.25
& \$2.19 \\

\midrule

NavJev
& \textbf{27.0}
& \textbf{22.4}
& 35.0
& \textbf{43.5}
& \textbf{0.53 $\pm$ 0.11}
& \textbf{0.65 $\pm$ 0.11}
& \textbf{\$0.06} \\

\bottomrule
\end{tabular}%
}
\end{table}

\begin{figure}[t]
    \centering
    \includegraphics[width=0.8\columnwidth]{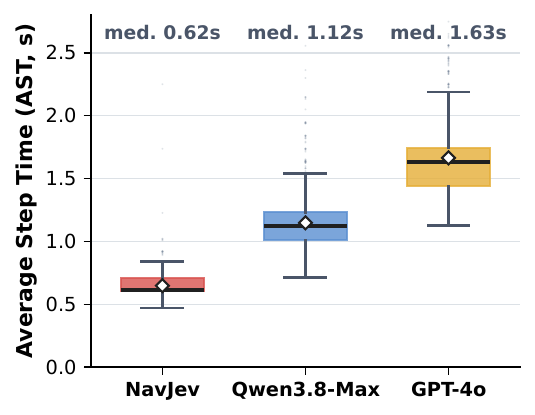}
    \caption{Distribution of per-episode average step time (AST) for different decision models under the same ACVC--DASM representation on the R2R-CE 100-episode split.}
    \label{fig:decision_model}
\end{figure}

\begin{figure*}[t]
    \centering
    \includegraphics[width=\textwidth]{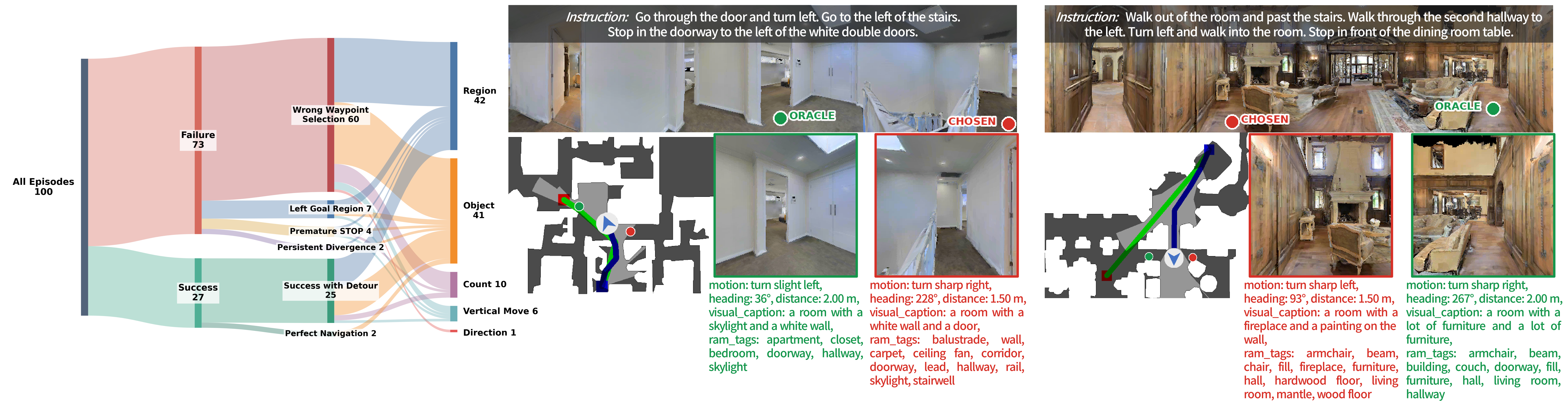}
    \caption{Failure analysis of NavJev on the R2R-CE 100-episode split. Left: distribution of navigation outcomes and failure modes, where incorrect waypoint selection accounts for most failures. Middle and right: representative failure cases comparing the waypoint selected by NavJev (red) with the oracle waypoint (green). When instructions require precise spatial-relation understanding or different viewpoints exhibit similar visual semantics, NavJev may select a plausible but incorrect direction.}
    \label{fig:fail}
\end{figure*}

\subsection{Inference Efficiency}

A central objective of NavJev is to reduce the online inference overhead caused by repeated large-model reasoning. Fig.~\ref{fig:timing_comparison} compares the average per-step inference latency of NavJev with representative zero-shot VLN methods and further analyzes the computational cost of the proposed pipeline.
As shown in Fig.~\ref{fig:timing_comparison}(a), NavJev requires only 0.65 $\pm$ 0.11\,s per navigation step, whereas Open-Nav, SmartWay, and P2DNav require 27.00 $\pm$ 3.64\,s, 5.30 $\pm$ 0.67\,s, and 4.92 $\pm$ 2.69\,s, respectively, corresponding to approximately 41.5$\times$, 8.2$\times$, and 7.6$\times$ speedups. The substantial reduction in per-step latency shows that compact action-centric representations and typed action selection can effectively reduce the computational overhead of repeated multimodal reasoning.

Fig.~\ref{fig:timing_comparison}(b) further presents the wall-clock latency breakdown of NavJev. The complete pipeline requires approximately 647.9\,ms per step, of which Jev decision making accounts for 534.6\,ms (82.5\%). In comparison, BLIP--RAM perception requires only 72.4\,ms (11.2\%), waypoint prediction 25.1\,ms (3.9\%), and environment interaction 14.9\,ms (2.3\%). These results show that the local perception pipeline introduces limited computational overhead, while the overall latency is dominated by decision inference. By replacing repeated autoregressive multimodal generation with compact ACVC--DASM representations and typed decisions, NavJev enables substantially faster closed-loop navigation.

We further compare NavJev with representative task-fine-tuned MLLM-based VLN methods in Table~\ref{tab:official_efficiency_compact}. Unlike NavJev, which selects waypoint-level macro-actions, these methods typically predict low-level discrete actions at each navigation step and therefore require substantially more decision steps to complete an episode. Although their per-step latency can be lower, NavJev remains within the same order of magnitude and is comparable to JanusVLN, while requiring only 9.3 steps and 6.03\,s per episode on average, substantially reducing the overall navigation time. NavJev also uses only 4.46\,GiB of peak GPU memory, considerably less than the compared methods. These results show that NavJev maintains low per-step inference overhead while substantially reducing end-to-end navigation time and computational resource consumption through fewer decision steps.

\subsection{Decision Model Analysis}

To analyze the effect of the decision model, we replace Jev with GPT-4o and Qwen3.8-Max while keeping the same ACVC--DASM representation and navigation pipeline. Reasoning is disabled for both autoregressive models, so all three models receive the same structured textual state and only select the next navigation action. Table~\ref{tab:backend_comparison} reports navigation performance, latency, and API cost, while Fig.~\ref{fig:decision_model} shows the distribution of per-step inference time.
NavJev achieves 27.0\% SR, outperforming Qwen3.8-Max and GPT-4o by 2.0\% and 13.0\%, respectively. It also reaches 22.4\% SPL and 43.5 nDTW, compared with 16.0\% SPL and 34.2 nDTW for Qwen3.8-Max, demonstrating better success, path efficiency, and trajectory fidelity.

\begin{figure*}[t]
    \centering
    \includegraphics[width=0.98\textwidth]{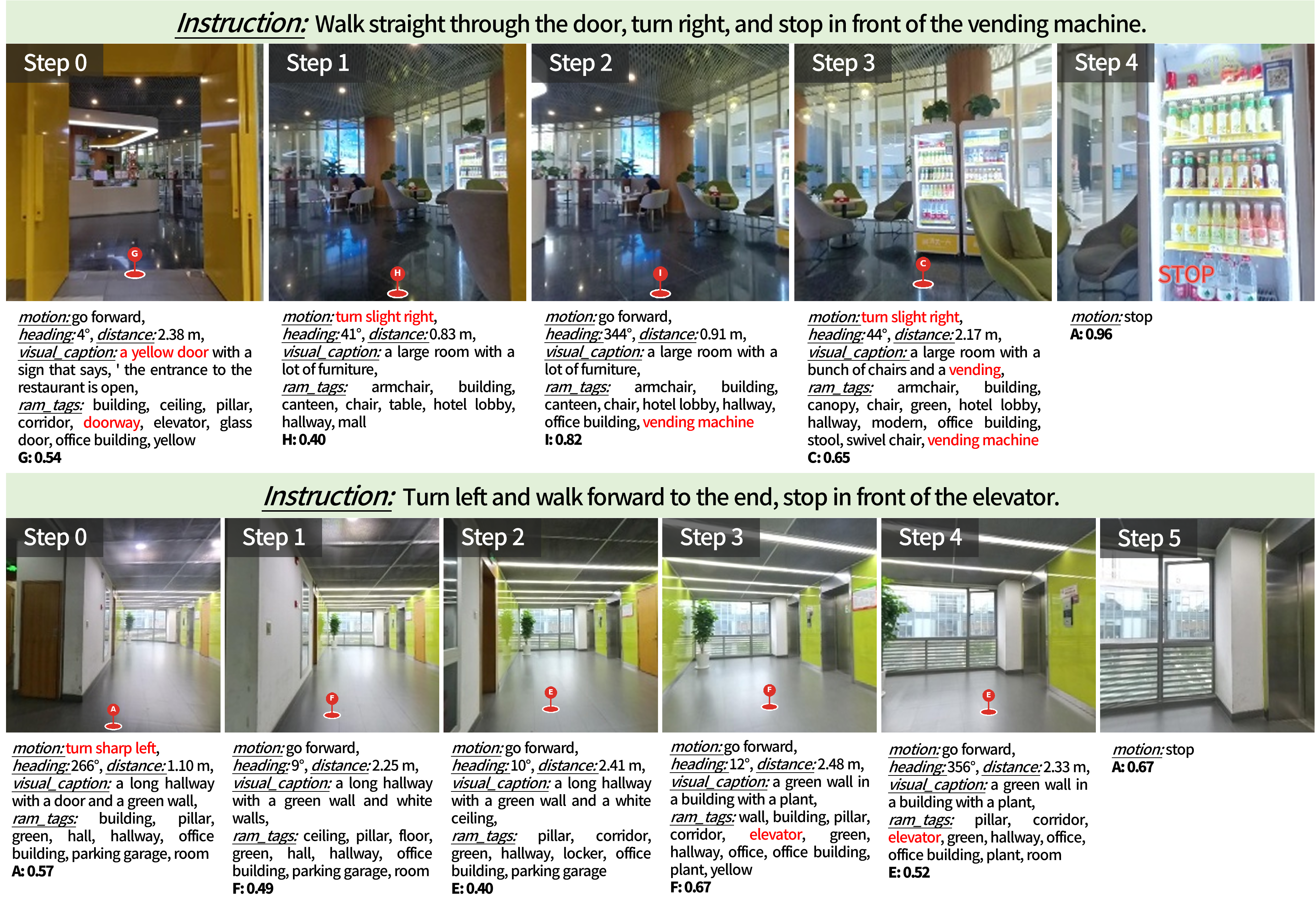}
    \caption{Real-world navigation visualization of NavJev. The agent follows natural-language instructions through sequential waypoint decisions and successfully stops at the referred vending machine and elevator. Red annotations indicate key selected actions and their corresponding action-semantic cues.}
    \label{fig:realworld_cases}
\end{figure*}

\begin{table}[t]
\centering
\caption{Component ablation of ACVC and DASM on the R2R-CE 100-episode split, where BLIP and RAM are components of ACVC.}
\label{tab:acvc_ablation}

\setlength{\tabcolsep}{2.8pt}
\resizebox{0.95\columnwidth}{!}{%
\begin{tabular}{ccc|cccccc}
\toprule
\textbf{BLIP} &
\textbf{RAM} &
\textbf{DASM} &
\textbf{TL}$\downarrow$ &
\textbf{NE}$\downarrow$ &
\textbf{OSR}$\uparrow$ &
\textbf{SR}$\uparrow$ &
\textbf{SPL}$\uparrow$ &
\textbf{nDTW}$\uparrow$ \\
\midrule

\cmark & \xmark & \xmark
& 12.10
& 8.80
& 33.0
& 19.0
& 14.4
& 36.2 \\

\cmark & \cmark & \xmark
& 15.83
& 8.13
& 34.0
& 21.0
& 16.1
& 35.0 \\

\cmark & \cmark & \cmark
& \textbf{11.58}
& \textbf{7.48}
& \textbf{35.0}
& \textbf{27.0}
& \textbf{22.4}
& \textbf{43.5} \\

\bottomrule
\end{tabular}%
}
\end{table}

NavJev also provides clear efficiency gains. Its decision latency is only 0.53\,s, compared with 1.03\,s for Qwen3.8-Max and 1.54\,s for GPT-4o, corresponding to 1.94$\times$ and 2.91$\times$ speedups. The total API cost over 100 episodes is only \$0.06, compared with \$2.19 and \$3.67, reducing cost by approximately 97.3\% and 98.4\%. These results show that typed decisions provide a stronger balance of navigation performance, latency, and cost than multimodal autoregressive decision making.

\subsection{Ablation Study}

We conduct ablation studies on the visual and semantic components of NavJev in Table~\ref{tab:acvc_ablation}. Using BLIP captions alone achieves 19.0\% SR and 14.4\% SPL. Adding RAM semantic tags improves SR to 21.0\% and SPL to 16.1\%, indicating that richer semantic information benefits navigation decisions.

Further introducing DASM yields more substantial improvements, increasing SR from 21.0\% to 27.0\% and SPL from 16.1\% to 22.4\%, corresponding to gains of 6.0\% and 6.3\%, respectively. DASM not only filters shared semantics to emphasize differences among candidate actions, but also preserves action-specific semantic evidence across navigation steps. These results demonstrate the benefit of combining discriminative semantic filtering with action-semantic memory for navigation decision making.

\subsection{Failure Analysis}
\label{sec:failure}

To better understand the limitations of NavJev, we analyze the failure modes of all 100 R2R-CE evaluation episodes in Fig.~\ref{fig:fail}. Among the 73 failed episodes, 60 failures (82.2\%) result from incorrect waypoint selection, making it the dominant error source. This suggests that the main bottleneck lies in distinguishing local candidate directions rather than termination or recovery behavior.

The qualitative examples in Fig.~\ref{fig:fail} illustrate this issue. In both cases, the selected and oracle directions contain highly similar scene cues with limited discriminative semantics. Although ACVC and DASM provide compact, discriminative descriptions, the correct action may still depend on fine-grained spatial relations in the instruction, such as \emph{to the left of the stairs} or \emph{the second hallway to the left}. When multiple directions are visually plausible and difficult to distinguish semantically, NavJev may select an incorrect waypoint and deviate from the intended route. These cases indicate that stronger fine-grained spatial grounding and relational reasoning remain important for further improving NavJev.

\begin{table}[t]
\centering
\caption{Real-world evaluation of NavJev in office and caf\'e environments. We compare NavJev with Qwen3.8-Max under the same waypoint candidates, ACVC perception, and DASM representation. OSR and SR are computed over 10 tasks per scene and averaged over all 20 tasks.}
\label{tab:realworld_navjev}

\setlength{\tabcolsep}{3.8pt}
\resizebox{\columnwidth}{!}{%
\begin{tabular}{lcccccccc}
\toprule
\multirow{2}{*}{\textbf{Decision Model}} &
\multicolumn{2}{c}{\textbf{Scene 1 (Office)}} &
\multicolumn{2}{c}{\textbf{Scene 2 (Caf\'e)}} &
\multicolumn{2}{c}{\textbf{Average}} &
\multirow{2}{*}{\shortstack{\textbf{Decision}\\\textbf{Latency (s)}$\downarrow$}} &
\multirow{2}{*}{\shortstack{\textbf{Average Step}\\\textbf{Time (s)}$\downarrow$}} \\

\cmidrule(lr){2-3}
\cmidrule(lr){4-5}
\cmidrule(lr){6-7}

& \textbf{OSR}$\uparrow$
& \textbf{SR}$\uparrow$
& \textbf{OSR}$\uparrow$
& \textbf{SR}$\uparrow$
& \textbf{OSR}$\uparrow$
& \textbf{SR}$\uparrow$
& & \\

\midrule

Qwen3.8-Max
& 30.0 & 30.0
& 40.0 & 40.0
& 35.0 & 35.0
& 1.10 $\pm$ 0.30
& 1.23 $\pm$ 0.31 \\

NavJev
& \textbf{50.0} & \textbf{40.0}
& \textbf{70.0} & \textbf{60.0}
& \textbf{60.0} & \textbf{50.0}
& \textbf{0.65 $\pm$ 0.51}
& \textbf{0.79 $\pm$ 0.51} \\

\bottomrule
\end{tabular}%
}
\end{table}

\subsection{Real-World Experiments}
\label{sec:real}

We further evaluate NavJev in two real-world indoor environments, including an office and a caf\'e, with 10 navigation tasks in each scene. NavJev and Qwen3.8-Max use the same waypoint candidates, ACVC perception, and DASM representation to ensure a fair comparison. Fig.~\ref{fig:realworld_cases} visualizes representative NavJev navigation trajectories together with the corresponding input representations and output decisions, while Table~\ref{tab:realworld_navjev} reports quantitative navigation performance and server-side inference efficiency. As shown in Table~\ref{tab:realworld_navjev}, NavJev achieves an average OSR of 60.0\% and SR of 50.0\%, whereas Qwen3.8-Max obtains 35.0\% for both metrics. The improvement is consistently observed in both environments, demonstrating the effectiveness of the proposed decision paradigm in real-world navigation.

NavJev also demonstrates clear efficiency advantages in real-world deployment. Its decision latency decreases from 1.10\,s to 0.65\,s, while the average step time decreases from 1.23\,s to 0.79\,s, corresponding to approximately 1.69$\times$ and 1.56$\times$ speedups, respectively. Together with the qualitative visualizations in Fig.~\ref{fig:realworld_cases}, these results indicate that the efficiency advantage of typed decision making transfers from simulation to physical navigation while maintaining effective and reliable navigation performance.
\section{Conclusion}
\label{sec:conclusion}

In this work, we presented NavJev, an efficient zero-shot Vision-and-Language Navigation framework that replaces repeated multimodal generation with compact action-centric representations and typed decision making. NavJev combines Action-Centric Visual Compression (ACVC), Discriminative Action-Semantic Memory (DASM), and Jev, where ACVC transforms visual observations into compact textual action representations, DASM filters shared semantics and retains action-specific evidence across navigation steps, and Jev performs lightweight typed action selection.
Experiments on R2R-CE show that NavJev achieves 27.0\% SR and 22.4\% SPL with only 0.65\,s per navigation step, while substantially reducing inference latency and cost compared with representative MLLM-based VLN methods. Real-world experiments further demonstrate that these efficiency advantages transfer to physical navigation. Overall, the results show that compact action-centric representations, discriminative action-semantic memory, and typed decisions provide an effective solution for efficient embodied navigation without repeated multimodal reasoning.

{\small
\bibliographystyle{ieeenat_fullname}
\bibliography{main}

@String(CVPR= {IEEE Conf. Comput. Vis. Pattern Recog.})

@String(ICCV= {Int. Conf. Comput. Vis.})

@String(AAAI = {AAAI})

@String(CVPRW= {IEEE Conf. Comput. Vis. Pattern Recog. Worksh.})

@String(CVPR  = {CVPR})

@String(ICCV  = {ICCV})

@String(CVPRW= {CVPRW})

@inproceedings{anderson2018r2r,
  title={Vision-and-language navigation: Interpreting visually-grounded navigation instructions in real environments},
  author={Anderson, Peter and Wu, Qi and Teney, Damien and Bruce, Jake and Johnson, Mark and S{\"u}nderhauf, Niko and Reid, Ian and Gould, Stephen and Van Den Hengel, Anton},
  booktitle={2018 IEEE/CVF conference on computer vision and pattern recognition},
  pages={3674--3683},
  year={2018},
  organization={IEEE}
}

@inproceedings{krantz2020vlnce,
  title={Beyond the nav-graph: Vision-and-language navigation in continuous environments},
  author={Krantz, Jacob and Wijmans, Erik and Majumdar, Arjun and Batra, Dhruv and Lee, Stefan},
  booktitle={European Conference on Computer Vision},
  pages={104--120},
  year={2020},
  organization={Springer}
}

@inproceedings{hong2021vlnbert,
  title={Vln bert: A recurrent vision-and-language bert for navigation},
  author={Hong, Yicong and Wu, Qi and Qi, Yuankai and Rodriguez-Opazo, Cristian and Gould, Stephen},
  booktitle={Proceedings of the IEEE/CVF conference on Computer Vision and Pattern Recognition},
  pages={1643--1653},
  year={2021}
}

@inproceedings{hao2020prevalent,
  title={Towards learning a generic agent for vision-and-language navigation via pre-training},
  author={Hao, Weituo and Li, Chunyuan and Li, Xiujun and Carin, Lawrence and Gao, Jianfeng},
  booktitle={2020 IEEE/CVF Conference on Computer Vision and Pattern Recognition (CVPR)},
  pages={13134--13143},
  year={2020},
  organization={IEEE}
}

@article{chen2021hamt,
  title={History aware multimodal transformer for vision-and-language navigation},
  author={Chen, Shizhe and Guhur, Pierre-Louis and Schmid, Cordelia and Laptev, Ivan},
  journal={Advances in neural information processing systems},
  volume={34},
  pages={5834--5847},
  year={2021}
}

@inproceedings{chen2022duet,
  title={Think global, act local: Dual-scale graph transformer for vision-and-language navigation},
  author={Chen, Shizhe and Guhur, Pierre-Louis and Tapaswi, Makarand and Schmid, Cordelia and Laptev, Ivan},
  booktitle={2022 IEEE/CVF Conference on Computer Vision and Pattern Recognition (CVPR)},
  pages={16516--16526},
  year={2022},
  organization={IEEE}
}

@inproceedings{wang2023scalevln,
  title={Scaling data generation in vision-and-language navigation},
  author={Wang, Zun and Li, Jialu and Hong, Yicong and Wang, Yi and Wu, Qi and Bansal, Mohit and Gould, Stephen and Tan, Hao and Qiao, Yu},
  booktitle={2023 IEEE/CVF International Conference on Computer Vision (ICCV)},
  pages={11975--11986},
  year={2023},
  organization={IEEE}
}

@article{an2023bevbert,
  title={BEVBert: Multimodal Map Pre-training for Language-guided Navigation},
  author={An, Dong and Qi, Yuankai and Li, Yangguang and Huang, Yan and Wang, Liang and Tan, Tieniu and Shao, Jing},
  journal={Proceedings of the IEEE/CVF International Conference on Computer Vision},
  year={2023}
}

@article{an2024etpnav,
  title={Etpnav: Evolving topological planning for vision-language navigation in continuous environments},
  author={An, Dong and Wang, Hanqing and Wang, Wenguan and Wang, Zun and Huang, Yan and He, Keji and Wang, Liang},
  journal={IEEE Transactions on Pattern Analysis and Machine Intelligence},
  volume={47},
  number={7},
  pages={5130--5145},
  year={2024},
  publisher={IEEE}
}

@article{he2025multilevel,
  title={A multilevel attention network with sub-instructions for continuous vision-and-language navigation: Z. He et al.},
  author={He, Zongtao and Wang, Liuyi and Li, Shu and Yan, Qingqing and Liu, Chengju and Chen, Qijun},
  journal={Applied Intelligence},
  volume={55},
  number={7},
  pages={657},
  year={2025},
  publisher={Springer}
}

@inproceedings{mee,
  title={Multimodal evolutionary encoder for continuous vision-language navigation},
  author={He, Zongtao and Wang, Liuyi and Chen, Lu and Li, Shu and Yan, Qingqing and Liu, Chengju and Chen, Qijun},
  booktitle={2024 IEEE/RSJ International Conference on Intelligent Robots and Systems (IROS)},
  pages={1443--1450},
  year={2024},
  organization={IEEE}
}

@inproceedings{wang2024goat,
  title={Vision-and-language navigation via causal learning},
  author={Wang, Liuyi and He, Zongtao and Dang, Ronghao and Shen, Mengjiao and Liu, Chengju and Chen, Qijun},
  booktitle={2024 IEEE/CVF Conference on Computer Vision and Pattern Recognition (CVPR)},
  pages={13139--13150},
  year={2024},
  organization={IEEE}
}

@article{zhang2024uninavid,
    title={Uni-NaVid: A Video-based Vision-Language-Action Model for Unifying Embodied Navigation Tasks},
    author={Zhang, Jiazhao and Wang, Kunyu and Wang, Shaoan and Li, Minghan and Liu, Haoran and Wei, Songlin and Wang, Zhongyuan and Zhang, Zhizheng and Wang, He},
    journal={Robotics: Science and Systems},
    year={2025}
}

@inproceedings{wei2025streamvln,
  title={Streamvln: Streaming vision-and-language navigation via slowfast context modeling},
  author={Wei, Meng and Wan, Chenyang and Yu, Xiqian and Wang, Tai and Yang, Yuqiang and Mao, Xiaohan and Zhu, Chenming and Cai, Wenzhe and Wang, Hanqing and Chen, Yilun and others},
  booktitle={2026 IEEE International Conference on Robotics and Automation (ICRA)},
  year={2026},
  organization={IEEE}
}

@inproceedings{zeng2025janusvln,
  title={Janusvln: Decoupling semantics and spatiality with dual implicit memory for vision-language navigation},
  author={Zeng, Shuang and Qi, Dekang and Chang, Xinyuan and Xiong, Feng and Xie, Shichao and Wu, Xiaolong and Liang, Shiyi and Xu, Mu and Wei, Xing},
  booktitle={International Conference on Learning Representations},
  volume={2026},
  pages={33001--33026},
  year={2026}
}

@article{sheng2026p2dnav,
  title={P2DNav: Panorama-to-Downview Reasoning for Zero-shot Vision-and-Language Navigation},
  author={Sheng, Kai and Wang, Liuyi and Dai, Haojie and Li, Jinlong and Qin, Yongrui and He, Zongtao and Liu, Chengju and Chen, Qijun},
  journal={arXiv preprint arXiv:2605.19634},
  year={2026}
}

@inproceedings{long2023discuss,
  title={Discuss before moving: Visual language navigation via multi-expert discussions},
  author={Long, Yuxing and Li, Xiaoqi and Cai, Wenzhe and Dong, Hao},
  booktitle={2024 IEEE International Conference on Robotics and Automation (ICRA)},
  pages={17380--17387},
  year={2024},
  organization={IEEE}
}

@article{zhang2024navid,
        title={NaVid: Video-based VLM Plans the Next Step for Vision-and-Language Navigation},
        author={Zhang, Jiazhao and Wang, Kunyu and Xu, Rongtao and Zhou, Gengze and Hong, Yicong and Fang, Xiaomeng and Wu, Qi and Zhang, Zhizheng and Wang, He},
        journal={Robotics: Science and Systems},
        year={2024}
      }

@inproceedings{zhou2024navgpt,
  title={Navgpt: Explicit reasoning in vision-and-language navigation with large language models},
  author={Zhou, Gengze and Hong, Yicong and Wu, Qi},
  booktitle={Proceedings of the AAAI Conference on Artificial Intelligence},
  volume={38},
  number={7},
  pages={7641--7649},
  year={2024}
}

@inproceedings{chen2024mapgpt,
  title={Mapgpt: Map-guided prompting with adaptive path planning for vision-and-language navigation},
  author={Chen, Jiaqi and Lin, Bingqian and Xu, Ran and Chai, Zhenhua and Liang, Xiaodan and Wong, Kwan-Yee},
  booktitle={Proceedings of the 62nd Annual Meeting of the Association for Computational Linguistics (Volume 1: Long Papers)},
  pages={9796--9810},
  year={2024}
}

@inproceedings{qiao2025opennav,
  title={Open-nav: Exploring zero-shot vision-and-language navigation in continuous environment with open-source llms},
  author={Qiao, Yanyuan and Lyu, Wenqi and Wang, Hui and Wang, Zixu and Li, Zerui and Zhang, Yuan and Tan, Mingkui and Wu, Qi},
  booktitle={2025 IEEE International Conference on Robotics and Automation (ICRA)},
  pages={6710--6717},
  year={2025},
  organization={IEEE}
}

@article{chen2025constraint,
  title={Constraint-aware zero-shot vision-language navigation in continuous environments},
  author={Chen, Kehan and An, Dong and Huang, Yan and Xu, Rongtao and Su, Yifei and Ling, Yonggen and Reid, Ian and Wang, Liang},
  journal={IEEE Transactions on Pattern Analysis and Machine Intelligence},
  year={2025},
  publisher={IEEE}
}

@inproceedings{shi2025smartway,
  title={Smartway: Enhanced waypoint prediction and backtracking for zero-shot vision-and-language navigation},
  author={Shi, Xiangyu and Li, Zerui and Lyu, Wenqi and Xia, Jiatong and Dayoub, Feras and Qiao, Yanyuan and Wu, Qi},
  booktitle={2025 IEEE/RSJ International Conference on Intelligent Robots and Systems (IROS)},
  pages={16923--16930},
  year={2025},
  organization={IEEE}
}

@inproceedings{chen2025aoplanner,
  title={Affordances-oriented planning using foundation models for continuous vision-language navigation},
  author={Chen, Jiaqi and Lin, Bingqian and Liu, Xinmin and Ma, Lin and Liang, Xiaodan and Wong, Kwan-Yee K},
  booktitle={Proceedings of the AAAI Conference on Artificial Intelligence},
  volume={39},
  number={22},
  pages={23568--23576},
  year={2025}
}

@article{wang2025clash,
  title={Clash: Collaborative large-small hierarchical framework for continuous vision-and-language navigation},
  author={Wang, Liuyi and He, Zongtao and Li, Jinlong and Xia, Ruihao and Hu, Mengxian and Yao, Chenpeng and Liu, Chengju and Tang, Yang and Chen, Qijun},
  journal={arXiv preprint arXiv:2512.10360},
  year={2025}
}

@InProceedings{li2022blip,
  title = 	 {{BLIP}: Bootstrapping Language-Image Pre-training for Unified Vision-Language Understanding and Generation},
  author =       {Li, Junnan and Li, Dongxu and Xiong, Caiming and Hoi, Steven},
  booktitle = 	 {Proceedings of the 39th International Conference on Machine Learning},
  pages = 	 {12888--12900},
  year = 	 {2022},
  volume = 	 {162}
}

@INPROCEEDINGS{zhang2024ram,
  author={Zhang, Youcai and Huang, Xinyu and Ma, Jinyu and Li, Zhaoyang and Luo, Zhaochuan and Xie, Yanchun and Qin, Yuzhuo and Luo, Tong and Li, Yaqian and Liu, Shilong and Guo, Yandong and Zhang, Lei},
  booktitle={2024 IEEE/CVF Conference on Computer Vision and Pattern Recognition Workshops (CVPRW)}, 
  title={Recognize Anything: A Strong Image Tagging Model}, 
  year={2024},
  volume={},
  number={},
  pages={1724-1732},
  doi={10.1109/CVPRW63382.2024.00179}
}

@misc{oquab2024dinov2,
  title={DINOv2: Learning Robust Visual Features without Supervision},
  author={Oquab, Maxime and Darcet, Timothée and Moutakanni, Theo and Vo, Huy V. and Szafraniec, Marc and Khalidov, Vasil and Fernandez, Pierre and Haziza, Daniel and Massa, Francisco and El-Nouby, Alaaeldin and Howes, Russell and Huang, Po-Yao and Xu, Hu and Sharma, Vasu and Li, Shang-Wen and Galuba, Wojciech and Rabbat, Mike and Assran, Mido and Ballas, Nicolas and Synnaeve, Gabriel and Misra, Ishan and Jegou, Herve and Mairal, Julien and Labatut, Patrick and Joulin, Armand and Bojanowski, Piotr},
  journal={arXiv:2304.07193},
  year={2023}
}

@misc{typesafe2026jev,
  author       = {Diogo Almeida},
  title        = {Introducing System One Models \& Jev},
  year         = {2026},
  month        = sep,
  howpublished = {TypeSafe AI Blog},
  note         = {Published September 15, 2026; accessed September 26, 2026},
  url          = {https://typesafe.ai/blog/introducing-system-one-models-and-jev}
}

@article{bai2025qwen3,
  title={Qwen3-vl technical report},
  author={Bai, Shuai and Cai, Yuxuan and Chen, Ruizhe and Chen, Keqin and Chen, Xionghui and Cheng, Zesen and Deng, Lianghao and Ding, Wei and Gao, Chang and Ge, Chunjiang and others},
  journal={arXiv preprint arXiv:2511.21631},
  year={2025}
}

@article{achiam2023gpt,
  title={Gpt-4 technical report},
  author={Achiam, Josh and Adler, Steven and Agarwal, Sandhini and Ahmad, Lama and Akkaya, Ilge and Aleman, Florencia Leoni and Almeida, Diogo and Altenschmidt, Janko and Altman, Sam and Anadkat, Shyamal and others},
  journal={arXiv preprint arXiv:2303.08774},
  year={2023}
}

@article{wang2026comprehensive,
  title={A Comprehensive Survey and Systematic Real-World Evaluation of Embodied Vision-and-Language Navigation},
  author={Wang, Liuyi and Sheng, Kai and He, Zongtao and Li, Jinlong and Qin, Yongrui and Dai, Haojie and Wang, Xiangyi and Yang, Jingwei and Yan, Qingqing and Liu, Chengju and others},
  journal={IEEE Transactions on Automation Science and Engineering},
  year={2026},
  publisher={IEEE}
}

@inproceedings{wang2025rethinking,
  title={Rethinking the embodied gap in vision-and-language navigation: A holistic study of physical and visual disparities},
  author={Wang, Liuyi and Xia, Xinyuan and Zhao, Hui and Wang, Hanqing and Wang, Tai and Chen, Yilun and Liu, Chengju and Chen, Qijun and Pang, Jiangmiao},
  booktitle={2025 IEEE/CVF International Conference on Computer Vision (ICCV)},
  pages={9455--9465},
  year={2025},
  organization={IEEE}
}

@article{wang2026magic,
  title={Magic: Meta-ability guided interactive chain-of-distillation for effective-and-efficient vision-and-language navigation},
  author={Wang, Liuyi and He, Zongtao and Shen, Mengjiao and Yang, Jingwei and Liu, Chengju and Chen, Qijun},
  journal={IEEE Transactions on Pattern Analysis and Machine Intelligence},
  year={2026},
  publisher={IEEE}
}

@inproceedings{cheng2025navila,
        title={Navila: Legged robot vision-language-action model for navigation},
        author={Cheng, An-Chieh and Ji, Yandong and Yang, Zhaojing and Gongye, Zaitian and Zou, Xueyan and Kautz, Jan and B{\i}y{\i}k, Erdem and Yin, Hongxu and Liu, Sifei and Wang, Xiaolong},
        booktitle={RSS},
        year={2025}
}

@inproceedings{NavFoM,
 author = {Zhang, Jiazhao and Li, Anqi and Qi, Yunpeng and Li, Minghan and Liu, Jiahang and Wang, Shaoan and Liu, Haoran and Zhou, Gengze and Wu, Yuze and Li, Xingxing and Fan, Yuxin and Li, Wenjun and Chen, Zhibo and Gao, Fei and Wu, Qi and Zhang, Zhizheng and Wang, He},
 booktitle = {International Conference on Learning Representations},
 pages = {127293--127322},
 title = {Embodied Navigation Foundation Model},
 volume = {2026},
 year = {2026}
}
}

\clearpage
\end{document}